\documentclass[runningheads]{llncs}
\usepackage[T1]{fontenc}
\usepackage{graphicx}
\usepackage{booktabs}
\usepackage[misc]{ifsym}
\newcommand{\corr}{(\Letter)}
\usepackage{mwe}

\usepackage{pifont}
\usepackage{tabularx}

\newcolumntype{C}{>{\centering\arraybackslash}X}
\usepackage{multirow}
\usepackage{rotating}
\usepackage{float}
\usepackage{longtable,verbatim}
\usepackage{xcolor}
\usepackage{subcaption}
\usepackage{amssymb}
\usepackage{hyperref}
\newcommand{\repeatthanks}{\textsuperscript{\thefootnote}}

\begin{document}

\title{Fine-tune Smarter, Not Harder: \\Parameter-Efficient Fine-Tuning for \\Geospatial Foundation Models}

\titlerunning{Parameter-Efficient Fine-Tuning for Geospatial Foundation Models}


\author{Francesc Marti Escofet\,\inst{1}\textsuperscript{,}\thanks{Equal contribution.} \and
Benedikt Blumenstiel\,\inst{1}\textsuperscript{,}\repeatthanks \corr \and
Linus Scheibenreif\,\inst{2} \and
Paolo Fraccaro\,\inst{1} \and
Konrad Schindler\,\inst{2}}
\authorrunning{Marti Escofet, F., Blumenstiel, B., Scheibenreif, L., Fraccaro, P. et al.}

\institute{
IBM Research Europe \and
ETH Zurich
 \\\email{benedikt.blumenstiel@ibm.com}}

\tocauthor{Francesc Marti Escofet, Benedikt Blumenstiel, Linus Scheibenreif, Paolo Fraccaro, Konrad Schindler}
\toctitle{Fine-tune Smarter, Not Harder: Parameter-Efficient Fine-Tuning for Geospatial Foundation Models}

\maketitle              

\begin{abstract}
Earth observation (EO) is crucial for monitoring environmental changes, responding to disasters, and managing natural resources. In this context, foundation models facilitate remote sensing image analysis to retrieve relevant geoinformation accurately and efficiently. However, as these models grow in size, fine-tuning becomes increasingly challenging due to the associated computational resources and costs, limiting their accessibility and scalability. Furthermore, full fine-tuning can lead to forgetting pre-trained features and even degrade model generalization.
To address this, Parameter-Efficient Fine-Tuning (PEFT) techniques offer a promising solution. In this paper, we conduct extensive experiments with various foundation model architectures and PEFT techniques to evaluate their effectiveness on five different EO datasets. Our results provide a comprehensive comparison, offering insights into when and how PEFT methods support the adaptation of pre-trained geospatial models. We demonstrate that PEFT techniques match or even exceed full fine-tuning performance and enhance model generalisation to unseen geographic regions, while reducing training time and memory requirements. Additional experiments investigate the effect of architecture choices such as the decoder type or the use of metadata, suggesting UNet decoders and fine-tuning without metadata as the recommended configuration. We have integrated all evaluated foundation models and techniques into the open-source package TerraTorch to support quick, scalable, and cost-effective model adaptation.

\end{abstract}

\section{Introduction}

Earth observation (EO) is concerned with collecting information about the Earth using remote sensing, and it has become indispensable for monitoring environmental changes, managing natural resources, and enabling rapid responses to natural disasters~\cite{zhu2017deep,sen1floods112020}. The ever-growing volume and complexity of EO data pose substantial challenges for effective analysis and interpretation. Advances in deep learning have helped to mitigate these challenges by automating and improving the accuracy of EO data analysis across diverse applications, including flood detection, land use classification, and climate monitoring~\cite{zhu2017deep}.

Foundation models (FMs), which leverage self-supervised learning, have recently emerged as powerful tools capable of capturing general-purpose representations from data. Domain-specific geospatial foundation models (GeoFMs), like Clay~\cite{clay2024} or Prithvi~\cite{prithvi2023,prithvieo22024}, are pre-trained on large-scale satellite datasets. These models promise superior performance, improved generalization capabilities, and ease of use across geospatial tasks~\cite{geobench,prithvieo22024,terratorch2024}. For instance, recent studies have demonstrated that GeoFMs outperform traditional deep learning baselines on tasks such as land cover mapping and crop classification~\cite{prithvieo22024}.

However, as GeoFMs follow the general trend towards ever larger models~\cite{clay2024,prithvieo22024}, fine-tuning them becomes more expensive and requires substantial computational resources. This limitation restricts their broader application and adoption. 
Parameter-Efficient Fine-Tuning (PEFT) techniques provide a viable solution to this challenge. By updating only a small subset of the model parameters during training, PEFT significantly reduces computational demands and memory usage without sacrificing performance~\cite{hanparameter}. PEFT techniques have established themselves in natural language processing and computer vision~\cite{hanparameter}, but to the best of our knowledge, they have not yet been systematically investigated in the context of GeoFMs.

\begin{figure}[tbh]
    \centering
    \includegraphics[width=0.95\textwidth]{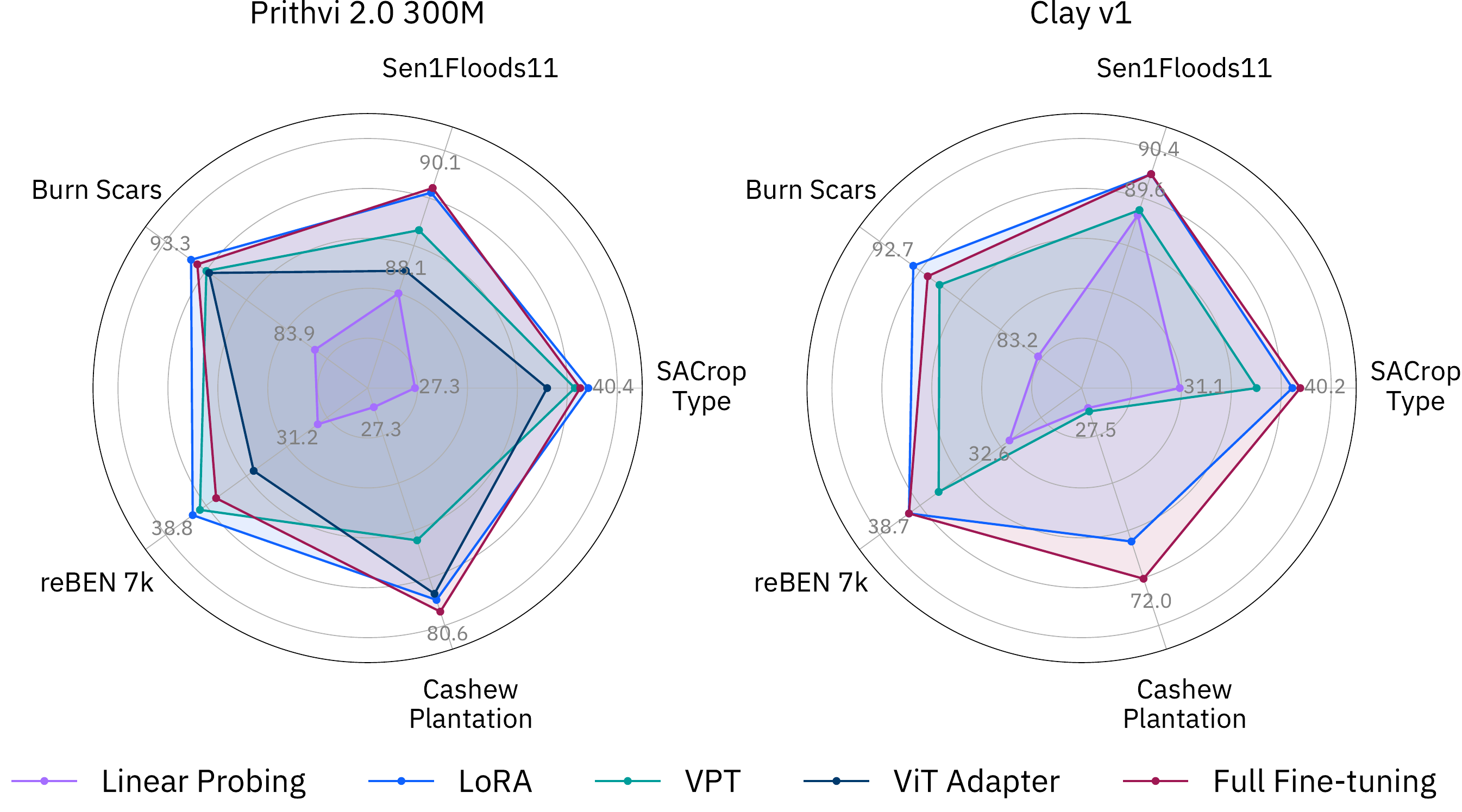}
    \caption{Comparison between PEFT techniques for Prithvi 2.0 300M (left) and Clay v1 (right) with linear decoders. The test mIoU (\%) $\uparrow$ is reported for all five datasets. Axes are scaled by min-max with dataset-specific buffers. The best and worst performance for each dataset is annotated.} 
    \label{fig:radar}
\end{figure}

Our work presents the first comprehensive evaluation of  PEFT techniques applied to GeoFMs. Specifically, we investigate multiple GeoFM architectures and PEFT methods across five distinct EO datasets, providing insights into their effectiveness and limitations. Our experimental findings, illustrated in Figure~\ref{fig:radar}, indicate that the Low-Rank Adaptation (LoRA) method matches or even exceeds the performance of full fine-tuning, while enhancing generalization to unseen geographic regions.

Our key contributions can be summarized as follows:
(1)~we perform the first extensive comparative analysis of PEFT methods for GeoFMs across diverse EO tasks, 
(2)~we explore the impact of architectural decisions, such as decoder choice and the inclusion of metadata, and 
(3)~we evaluate model generalization to unseen geographical regions and inputs, demonstrating the robustness of PEFT.
All evaluated models and techniques are integrated into the open-source package \texttt{TerraTorch}~\cite{terratorch2024} to promote the application of efficient model adaptation techniques in Earth observation.
The fine-tuning configs and code for our experiments are available at \url{https://github.com/IBM/peft-geofm}.

\section{Related Work}

Due to the growing volume of EO data available, machine learning techniques have become crucial to extract insights from these observations. A large body of work addresses the automated processing of EO data with deep learning techniques for a broad set of tasks, including, among others, environmental monitoring, land-cover mapping, and agricultural applications (see~\cite{zhu2017deep} for a review).

\paragraph{Foundation Models in EO.}
Foundation models are large neural network models pre-trained on vast datasets through self-supervised learning~\cite{bommasani2021opportunities}. They have led to significant progress in domains such as natural language processing~\cite{brown2020language} or computer vision~\cite{caron2021emerging}, and multiple foundation models for EO have recently been released~\cite{prithvi2023,xiong2024neural,decur2024}. 
They all leverage extensive collections of unlabelled remote sensing data to pre-train a generic data representation, using techniques such as contrastive learning~\cite{hadsell2006dimensionality} or masked autoencoding~\cite{he2022masked}.
E.g., Prithvi~1.0~\cite{prithvi2023} uses time-series of remote sensing data in conjunction with masked autoencoding to create an EO foundation model. Similarly, approaches like Clay~\cite{clay2024} and DOFA~\cite{xiong2024neural} leverage multi-modal remote sensing data with wavelength encodings to pre-train models for EO tasks. Contrastive GeoFM methods such as DeCUR~\cite{decur2024} achieve strong performance on some downstream tasks with relatively small model architectures. Overall, foundation models owe their performance to large model capacity (resp., size) and massive amounts of unlabeled data for pre-training. In the EO domain, recent models like Prithvi~2.0~\cite{prithvieo22024} demonstrate improved downstream performance by further scaling up both model size and data volume. 

\paragraph{PEFT Methods.}
EO foundation models are typically adapted for a concrete downstream task by fine-tuning all model parameters in a supervised fashion for the target task~\cite{prithvieo22024,decur2024}. With growing model size, such full fine-tuning incurs increasingly high computational costs and memory requirements. That bottleneck can be alleviated with PEFT methods, which drastically limit the number of fine-tuned parameters while still maintaining good downstream performance.
The core idea of PEFT is to remain efficient by making an informed choice about which parameters of the pre-trained model to update during fine-tuning~\cite{zaken2022bitfit}, or by introducing a small set of new parameters into the model~\cite{chenvision} or its input space~\cite{jia2022visual} and training only those.  
Several recent works introduce PEFT techniques specifically for EO data. For instance, methods like AiRs~\cite{hu2024airs} and TEA~\cite{hu2024tea} introduce adapters with additional residual connections for efficient fine-tuning of pre-trained transformers for EO. Similarly, UPetu~\cite{dong2024upetu} combines quantization with a learnable prompt to adapt pre-trained convolutional models to dense EO tasks. DEFLECT~\cite{deflect2025} further contributes to this direction by introducing a patch embedding layer (UPE) and an attention mechanism (uAtt) to adapt RGB-pretrained GeoFMs to multispectral images. Recent work also evaluates different PEFT methods on specific EO tasks, like the segmentation of winter-wheat fields~\cite{zahweh2023empirical}, or employs PEFT for unsupervised domain adaptation in EO~\cite{scheibenreif2024parameter}.

\noindent Unlike existing works, this work compares multiple PEFT methods across different multispectral EO foundation models and geospatial tasks. 

\section{Methods}
We first introduce the different PEFT methods as well as the GeoFMs models and decoders used in our experiments.
\begin{figure}[tb]
    \centering
     \begin{subfigure}[b]{0.28\textwidth}
         \centering
         \includegraphics[width=\textwidth]{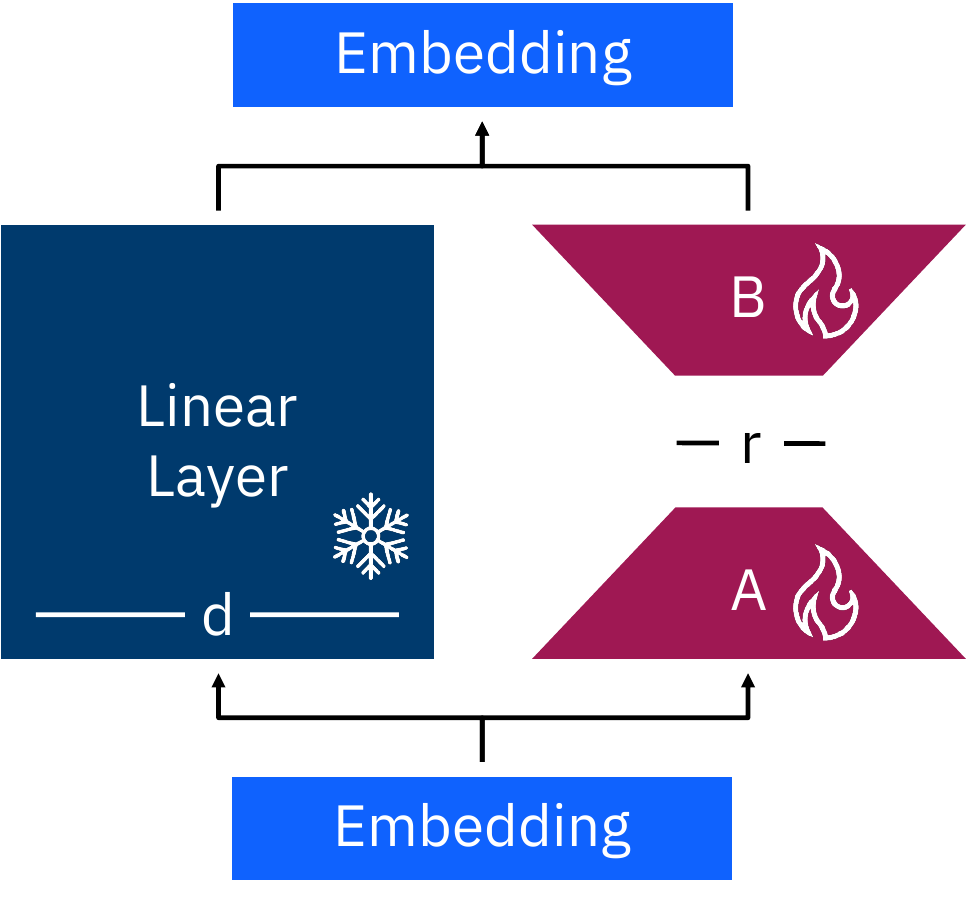}
         \caption{LoRA}
     \end{subfigure}
     \hfill
     \begin{subfigure}[b]{0.28\textwidth}
         \centering
         \includegraphics[width=\textwidth]{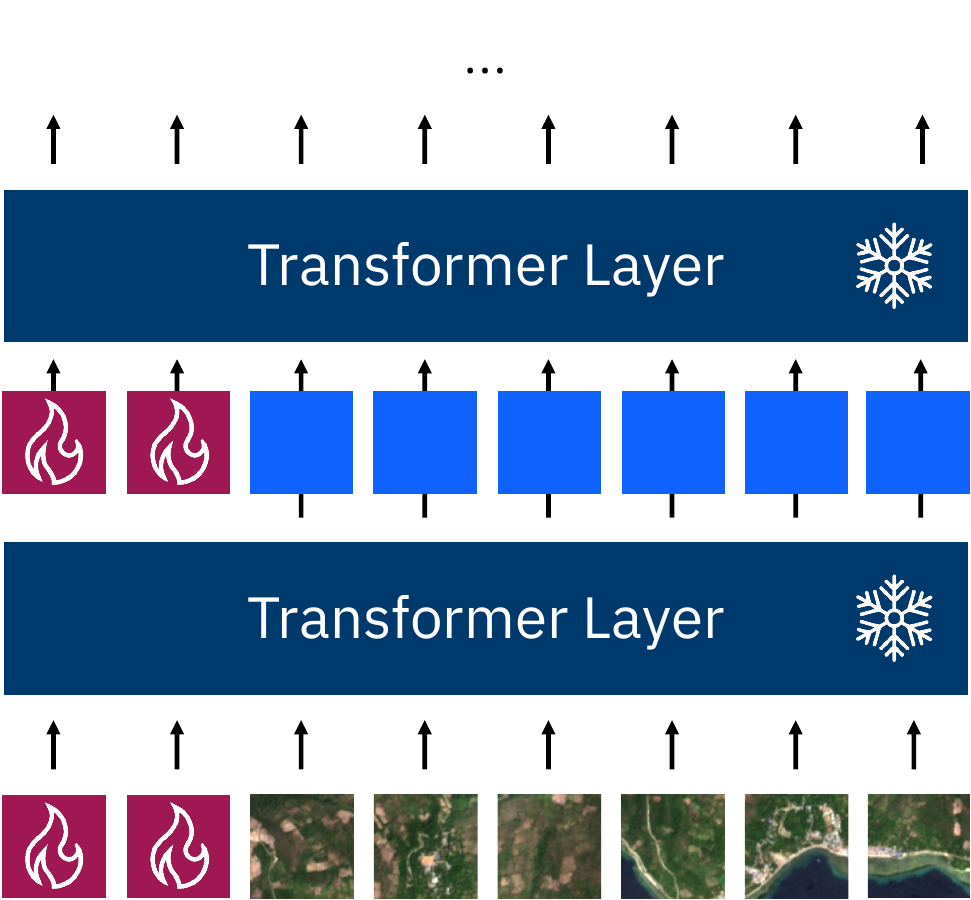}
         \caption{VPT}
     \end{subfigure}
     \hfill
     \begin{subfigure}[b]{0.28\textwidth}
         \centering
         \includegraphics[width=\textwidth]{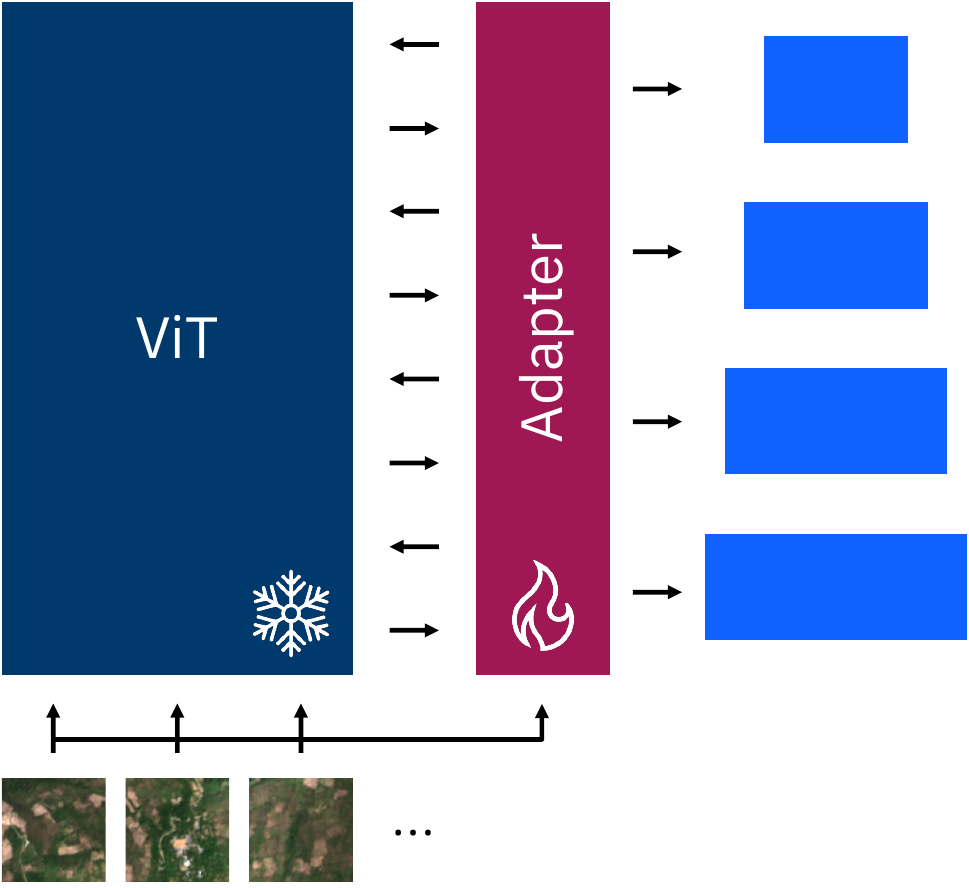}
         \caption{ViT Adapter}
     \end{subfigure}
    \caption{Architectures of PEFT techniques, visualized with embeddings (blue), frozen (dark blue), and learnable layers (red). LoRA introduces small weight matrices that modulate linear layers. VPT adds learnable prompt tokens to each layer of a ViT. The ViT Adapter is a smaller parallel network that interacts with multiple transformer layers and maps their activations to a feature hierarchy.}
    \label{fig:peft}
\end{figure}

\paragraph{PEFT Techniques.}
We combine three different PEFT methods with GeoFMs, see Figure~\ref{fig:peft}. Low-Rank Adaptation~\cite{hu2022lora} (LoRA) introduces low-rank matrices into pre-trained transformer models ($A$ and $B$ in Figure~\ref{fig:peft}). Fine-tuning the low-rank matrices reduces the amount of trainable parameters for a pre-trained weight matrix $W\in\mathbb{R}^{d\times d}$ from $d^2$ to $2\cdot d\cdot r$, where $r$ is the bottleneck dimension.

Instead of adapting the pre-trained model, visual prompt tuning~\cite{jia2022visual} (VPT) introduces additional trainable prompt parameters into the input sequence of the embedded image patches before the GeoFM processes them. A variant of that idea introduces learnable prompts in all transformer layers (VPT-Deep), the configuration we adopted here.

ViT Adapters~\cite{chenvision} address shortcomings of vision transformers (ViT)~\cite{vit2020} in dense prediction tasks by introducing spatial inductive biases. A small convolutional model extracts spatial features from the input image, which are then injected into the ViT through cross-attention modules. Eventually, the ViT adapters extract multi-scale features from the transformer layers which are combined for a final dense prediction. As ViT Adapters involve integrating convolutional branches, cross-attention modules, and multi-scale feature fusion, their implementation is non-trivial. Due to this significant engineering overhead, we only study them for the Prithvi family of models.

\begin{table}[tbh]
\centering
\setlength{\tabcolsep}{2pt}
\caption{Overview of the EO backbones and their pre-training characteristics.}
\label{tab:backbones}
\begin{tabularx}{\linewidth}{lCcCccC}
\toprule
Model & Archi. & \# Params. & Sensors & \# S-2 bands & \# Samples & Metadata \\
\midrule
DeCUR~\cite{decur2024} & ResNet50 & 25M & \mbox{S1, S2} & 13 & 1M & \ding{55} \\
Clay v1~\cite{clay2024} & ViT-B/8 & 92M & various & 10 & 70M & \ding{51} \\ 
Prithvi 1.0~\cite{prithvi2023} & ViT-B/16 & 86M & HLS & 6 & 250k $\times$ 3 & \ding{55}\\ 
Prithvi 2.0~\cite{prithvieo22024} & ViT-L/16 & 304M & HLS & 6 & 4.2M $\times$ 4 & \ding{55} / \ding{51}  \\
\bottomrule
\end{tabularx}
\end{table}

\paragraph{Foundation Models.}
We evaluate the different PEFT schemes with four GeoFMs and provide an overview in Table~\ref{tab:backbones}. DeCUR ~\cite{decur2024} is pre-trained with contrastive learning. Specifically, the Decoupling Common and Unique Representations (DeCUR) approach separates shared and modality-specific features in the embedding. The model utilises per-modality ResNet-50 backbones and was pre-trained on the SSL4EO-S12~\cite{wang2023ssl4eo} dataset, using all 13 Sentinel-2~L1C bands and two Sentinel-1~GRD bands.

Prithvi 1.0~\cite{prithvi2023} employs masked autoencoder (MAE)~\cite{he2022masked} pre-training and is based on a ViT-B backbone with a patch size of 16. It is trained on Harmonized Landsat and Sentinel-2 data exclusively from the USA, limited to six spectral bands. The data consists of short time series with three timestamps to support multi-temporal inputs naturally. Like its predecessor, Prithvi 2.0~\cite{prithvieo22024} employs MAE pre-training but uses a global dataset and includes larger backbones based on ViT-L/16 and ViT-H/14. While still training on HLS and covering six bands, optional temporal and location embeddings are introduced to enable better spatiotemporal reasoning. We evaluate the Prithvi 2.0 300M version, and do not use metadata unless specified otherwise.

Clay v1~\cite{clay2024} also pre-trains a ViT-B backbone, with a patch size of 8, by MAE augmented with DINO-style self-distillation~\cite{caron2021emerging}. It is trained on a diverse dataset that includes Landsat-8 and -9, Sentinel-1 and -2, NAIP, and LINZ imagery. It features dynamic patch embedding weights similar to DOFA~\cite{xiong2024neural}, adjusted according to the input spectral bands. The model integrates additional metadata like latitude/longitude, ground sampling distance (GSD), and acquisition time into the positional embeddings.

\paragraph{Decoder.}

The downstream performance of a GeoFM depends on the quality of the encoder features and the decoder's capability to leverage them. We evaluate four decoder architectures: a linear decoder, Fully Convolutional Network (FCN)~\cite{fcn2015}, UperNet~\cite{upernet2018}, and UNet~\cite{unet2015}. With its minimal complexity of a single transposed convolution without nonlinearities, the linear decoder serves as a way to evaluate the feature embeddings as directly as possible. 

While the linear decoder is well-suited for controlled evaluations of the encoder, real-world applications require decoders capable of exploiting features in context. The FCN decoder~\cite{fcn2015} reconstructs spatial details through a sequence of convolutional blocks, each composed of a transposed convolution for upsampling, followed by a 3$\times$3 convolution, Layer Normalization, and GeLU activation. 
The UperNet decoder~\cite{upernet2018} further improves upon this by combining multi-scale features through a Feature Pyramid Network (FPN)~\cite{lin2017feature} and capturing global context with a Pyramid Pooling Module (PPM)~\cite{zhao2017pyramid}. 
Lastly, the UNet decoder~\cite{unet2015} integrates multi-scale encoder features via skip connections, preserving detailed spatial information through decoder blocks consisting of interpolation, concatenation, and successive convolutional layers paired with Batch Normalization and ReLU activations.

The UperNet and UNet decoders require hierarchical, multi-scale features from multiple encoder layers. In our experiments, we use features from three intermediate encoder layers and the final one. While ResNet backbones naturally contain these hierarchical features, ViT embeddings require spatial upscaling. We therefore add learned upsampling layers that generate appropriate multi-scale features compatible with the UperNet and UNet decoders~\cite{terratorch2024}.

\section{Experimental Setup}

We have conducted extensive experiments to compare different PEFT techniques and GeoFMs. In the following, we describe the datasets used for evaluation and the fine-tuning protocol in detail.

\paragraph{Datasets.}
An overview of the data used for downstream evaluations is provided in Table~\ref{tab:datasets}. We restrict all experiments to multispectral optical inputs only.

\begin{table}[tbh]
\centering
\setlength{\tabcolsep}{2pt}
\caption{Overview of the evaluation datasets and their sample counts, including geographic hold-out sets (GHOS).}
\label{tab:datasets}
\begin{tabularx}{\linewidth}{lcccCCCCC}
\toprule
Dataset & Sensors & Patch size & Classes & \#\,Train & \#\,Val. & \#\,Test & \#\,GHOS \\
\midrule
Sen1Floods11~\cite{sen1floods112020} & S1, S2 & $512\times 512$ & 2 & 252 & 89 & 90 & 15 \\
Burn Scars~\cite{burnscars2023} & HLS & $512\times 512$ & 2 & 524 & 160 & 120 & – \\ 
reBEN 7k~\cite{clasen2024reben} & S1, S2 & $120\times 120$ & 19 & 4690 & 946 & 945 & 807 \\ 
m-Cashew plant.~\cite{cashew2021} & S2 & $256\times 256$ & 7 & 1350 & 400 & 50 & –  \\
m-SA Crop type~\cite{sacroptype} & S2 & $256\times 256$ & 10 & 3000 & 1000 & 1000 & –  \\
\bottomrule
\end{tabularx}
\end{table}

Sen1Floods11~\cite{sen1floods112020} includes Sentinel-1 (S1) GRD and Sentinel-2 (S2) L1C satellite images at 10m resolution covering eleven different flooding events on six different continents. The task is to segment surface water. In our evaluation, we use the 431 binary masks annotated by expert analysts. A hold-out set of 15 samples from Bolivia serves to test geographic generalization.

The Burn Scars dataset~\cite{burnscars2023} contains images of wildfire scars, gathered with the help of the Monitoring Trends in Burn Severity (MTBS) historical fire database. Annotations from the continental US, from 2018 to 2021, were co-registered with Harmonized Landsat Sentinel-2 (HLS) observations at 30m resolution. The 804 images include six spectral channels: Blue, Green, Red, NIR Narrow, SWIR 1, and SWIR 2.
The original Burn Scars dataset provides a training and validation split. We propose new splits that include a test set. To further exclude the possibility of data leakage, we create the splits with a 5\,km buffer between training and evaluation samples and ensure that close-by patches are assigned to the same split. We make the splits public at \url{https://github.com/IBM/peft-geofm} to ensure reproducibility.

reBEN~7k is a subset of the larger Refined BigEarthNet (reBEN)~\cite{clasen2024reben}, sampled to preserve class balance. reBEN has over 400k samples, each consisting of S1~GRD and S2~L2A patches as well as land-use/land-cover annotations with 19~classes. While reBEN supports both semantic segmentation and multi-label classification, we only use the semantic segmentation task.
Repeated fine-tuning on the full reBEN dataset would be computationally infeasible for our experiments. Therefore, we create a smaller subset following the example of BEN-ge~8k~\cite{mommert2023benge}. Compared to BEN-ge~8k, we derive our version from the latest edition of BigEarthNet~\cite{clasen2024reben} and introduce a geographic hold-out set (GHOS) for generalization experiments. Our subset is based on the land-cover multi-labels and designed to retain class diversity. To ensure fair representation of all classes, we draw 250 samples per class from the official training set and 50 samples per class for validation and testing, excluding Austria and Ireland. Then, up to 50 samples per class were drawn only from those countries' test sets to obtain an additional GHOS for testing model generalization to unseen geographic regions.

As further test cases, we include the m-Cashew Plantation~\cite{cashew2021} and m-SA~Crop Type~\cite{sacroptype} datasets provided by GEO-bench~\cite{geobench}. The former provides labeled images of cashew plantations across a 120 km$^2$ region in central Benin. Each pixel is assigned to one of seven classes, including \textit{well-managed plantation}, \textit{poorly-managed plantation}, and \textit{no plantation}, enabling detailed land-use analysis. SA Crop Type~\cite{sacroptype} contains images and crop type labels for South Africa, with each pixel classified into one of ten categories, such as \textit{weeds}, \textit{wheat}, or \textit{rooibos}. In both cases, we use the versions preprocessed by GEO-bench~\cite{geobench}.

\paragraph{Fine-tuning Procedure.}

To ensure robust and reproducible results, we follow established experimental protocols~\cite{geobench,prithvieo22024} involving hyperparameter optimization (HPO) and repeated runs. For each combination of model architecture, dataset, and PEFT method, we carry out Bayesian hyperparameter optimization with 16 trials to select the optimal learning rate, implemented using TerraTorch Iterate~\cite{iterate2025}. With the best learning rate, we run the fine-tuning five times for each configuration, starting from different random seeds, and report the average. 

Based on prior experiments, PEFT methods are configured as follows: For LoRA, adapters with bottleneck dimension $r {=} 16$ are added to the linear projections in the feed-forward layers, as well as to the query and value matrices of the attention layers of the GeoFMs. For VPT, we include 100 learnable prompts in all transformer layers (VPT-Deep). The specific parameter counts for each model and PEFT configuration are described in Table~\ref{tab:backbones_peft}.
VPT and LoRA increase the model parameters by only up to 2.4\%. ViT Adapter introduces significantly more parameters, increasing the model parameter count by up to 13.8\%.

\begin{table}[tb]
\centering
\setlength{\tabcolsep}{2pt}
\caption{Parameter counts for EO backbones and PEFT components. Percentages refer to the number of PEFT parameters relative to the encoder parameters.}
\label{tab:backbones_peft}
\begin{tabularx}{\linewidth}{lCCCC}
\toprule
Model & \# Encoder Params. & \# VPT Params. & \# LoRA Params. & \# ViT Adapt. Params. \\
\midrule
Clay v1~\cite{clay2024} & 92M & 0.9M (1.0\%) & 2M (2.2\%) & - \\ 
Prithvi 1.0 100M~\cite{prithvi2023} & 86M & 0.9M (1.1\%) & 2M (2.4\%) & 12M (13.8\%)\\ 
Prithvi 2.0 300M~\cite{prithvieo22024} & 304M & 2.5M (0.8\%) & 5.5M (1.8\%) & 20M (6.6\%)  \\
\bottomrule
\end{tabularx}
\end{table}

For all experiments, we use the AdamW optimizer with $\beta_1 {=} 0.9$, $\beta_2 {=} 0.999$ and ReduceLROnPlateau scheduler ($patience{=}4$ epochs, $factor{=}0.5$). The batch size was set to 8, except for reBEN-7k, where we increased it to 32 to account for the smaller input size and higher number of images. We find that a small batch size with a correspondingly higher number of gradient updates leads to faster convergence on relatively small EO datasets. All models were trained on a single NVIDIA A100 GPU for up to 100 epochs, applying early stopping after 15 epochs without improvement.
As far as possible, each GeoFM receives the Sentinel-2 bands it was pre-trained for: Ten bands for Clay, six for Prithvi, and all 13 L1C bands for DeCUR. If any bands are missing in a dataset (e.g., Burn Scars only has six bands), the patch embedding automatically adapts to the available subset via TerraTorch~\cite{terratorch2024}.
For reBEN-7k, the images are padded to $128\times128$ pixels with the \textit{reflect} method to ensure compatibility across all architectures.


\section{Experiments}

We conduct a series of experiments to evaluate the effectiveness of PEFT methods for GeoFMs. Our analyses cover three aspects: (1)~a comparison of PEFT techniques against full fine-tuning, (2)~an assessment of generalization across unseen geographic regions, and (3)~an evaluation of different decoder architectures.

\subsection{Parameter-Efficient Fine-Tuning}

Our experiments use a linear decoder to isolate the impact of PEFT techniques on geospatial foundation models. It ensures a controlled evaluation, as more complex decoders may obscure differences by significantly increasing model capacity.
Table~\ref{tab:peft} presents the results of PEFT experiments, including three baseline models: a randomly initialized UNet~\cite{unet2015} and a ViT-B/16 pre-trained on ImageNet-21k~\cite{vit2020}. DeCUR~\cite{decur2024}, a ResNet-based GeoFM, is also considered a baseline since the selected PEFT methods are designed for ViTs. We perform five training runs with varying random seeds and report average performances for readability. Standard deviations are provided in the supplementary material. In most cases, the standard deviation is below 0.6pp, with some exceptions for the Cashew Plantation dataset.

\begin{table}[tbh]
    \centering
    \caption{Test mIoU (\%) $\uparrow$ for the evaluated models and PEFT techniques, averaged over five runs. We highlight the best-performing combination in bold and underline the second-best one.}
    \label{tab:peft}
    \setlength{\tabcolsep}{2pt}
    \begin{tabularx}{\linewidth}{llCCCCCC}
    \toprule
    Model & Method & Sen1F11 & Burn Scars & reBEN 7k & Cashew Plant. & SACrop & Mean \\
    \midrule
    UNet (Rand.) & Full FT & \textbf{90.75} & 89.34 & 34.62 & \textbf{81.81} & 34.99 & 66.30 \\
    \midrule
    ViT (IN-21k) & Full FT & 89.19 & 92.31 & 36.15 & 80.05 & 37.98 & 67.14 \\
    \midrule
    \multirow{2}{*}{DeCUR} & Linear Prob. & 80.83 & 78.17 & 28.29 & 16.05 & 20.93 & 44.85 \\
     & Full FT & 86.87 & 89.48 & 36.05 & 79.59 & 34.21 & 65.24 \\
    \midrule
    \multirow{4}{*}{Clay v1} & Linear Prob. & 89.57 & 83.17 & 32.65 & 27.50 & 31.10 & 52.80 \\
    & VPT & 89.67 & 90.67 & 36.87 & 28.38 & 36.89 & 56.50 \\
     & LoRA & \underline{90.41} & 92.74 & \underline{38.67} & 62.29 & 39.64 & 64.75 \\
     & Full FT & \underline{90.41} & 91.58 & \underline{38.67} & 72.03 & \underline{40.22} & 66.58 \\
    \midrule
    \multirow{5}{*}{Prithvi 1.0 100M} & Linear Prob. & 88.78 & 83.23 & 27.07 & 25.33 & 26.06 & 50.10 \\
     & VPT & 89.03 & 85.17 & 29.98 & 29.24 & 29.62 & 52.61 \\
     & LoRA & 89.33 & 89.34 & 30.60 & 53.00 & 31.58 & 58.77 \\
     & ViT Adapter & 87.72 & 89.47 & 32.90 & 73.31 & 32.34 & 63.15 \\
     & Full FT & 89.02 & 89.31 & 31.82 & 77.41 & 32.59 & 64.03 \\
    \midrule
    \multirow{5}{*}{Prithvi 2.0 300M} & Linear Prob. & 88.08 & 83.90 & 31.25 & 27.29 & 27.26 & 51.55 \\
     & VPT & 89.31 & 92.16 & 38.40 & 62.00 & 39.33 & 64.24 \\
     & LoRA & 90.04 & \textbf{93.33} & \textbf{38.84} & 77.53 & \textbf{40.35} & \underline{68.02} \\
     & ViT Adapter & 88.52 & 91.95 & 35.14 & 75.92 & 37.24 & 65.75 \\
     & Full FT & 90.13 & \underline{92.85} & 37.42 & \underline{80.58} & 39.74 & \textbf{68.14} \\
    \bottomrule
    \end{tabularx}
\end{table}

\begin{figure}[tbh]
    \centering
    \includegraphics[width=0.9\linewidth]{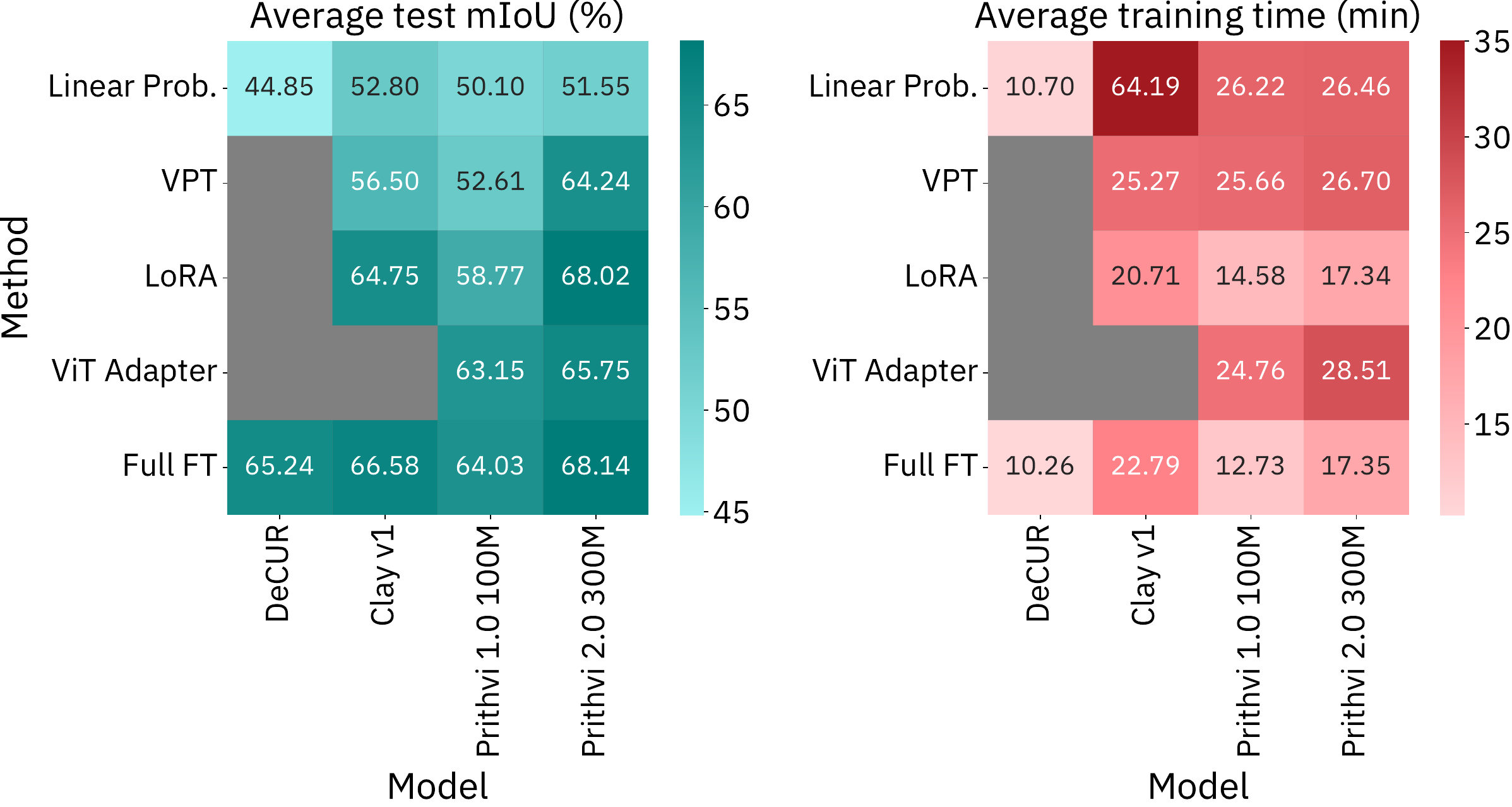}
    \caption{Test mIoU $\uparrow$ and training time $\downarrow$ averaged over five datasets with five runs each. We trained all models up to 100 epochs and used early stopping after 15 epochs, resulting in best efficiencies for full fine-tuning and LoRA.}
    \label{fig:time_epoch}
\end{figure}

While performance varies across datasets, Prithvi 2.0 300M is the best model overall, outperforming Clay v1 and the ImageNet-pre-trained ViT by +1pp. Notably, the randomly initialized UNet surpasses DeCUR and Prithvi 1.0. However, the superior UNet decoder may contribute to this advantage.
Prithvi 1.0 exhibits the lowest fine-tuning performance, likely due to its limited pre-training data. In contrast, Prithvi 2.0, trained on a larger dataset, achieves +4pp improvement, reinforcing the importance of scaling both data and model size~\cite{prithvieo22024}.
Among PEFT techniques, LoRA performs on par with, or better than, full fine-tuning for Clay and Prithvi 2.0. In contrast, VPT and ViT Adapter generally underperform full fine-tuning. On reBEN~7k, VPT and LoRA outperform full fine-tuning using Prithvi 2.0, while with the smaller Prithvi 1.0, ViT Adapter scores best. In general, PEFT methods are particularly performant when combined with large GeoFMs, like Prithvi 2.0.

Figure~\ref{fig:time_epoch} shows test performance compared with training time. LoRA and full fine-tuning require a similar training time for Prithvi 2.0, despite LoRA freezing most layers. This suggests that the fixed batch size for all experiments may limit LoRA's expected speedup. However, the lower memory footprint of LoRA makes it well-suited for GPUs with limited capacity and enables larger batch sizes for efficiency gains on larger datasets (see memory footprints in the supplementary material).
Training time is also strongly influenced by model architecture. Clay, which uses a patch size of 8, must process four times more tokens than Prithvi 1.0 and takes more than twice as long, despite their similar parameter counts. Prithvi 2.0, being three times larger than Clay, still trains faster.

An interesting finding is the unexpectedly long training time of linear probing shown in Figure~\ref{fig:time_epoch}, as the fitting of a less flexiblelinear function converges slower. Notably, Clay outperforms Prithvi 2.0 by 1pp in linear probing, but ranks lower in all other settings. 
However, this trend does not hold for larger decoders, which can model non-linear patterns. Additional experiments with a UNet decoder showed training times comparable to full fine-tuning. When combined with larger decoders, our results suggest that freezing the encoder is viable for simple tasks like water mapping but leads to performance drops in more complex tasks like crop type classification.

Overall, our results indicate that PEFT techniques offer an effective trade-off between full fine-tuning and frozen encoders, particularly when memory efficiency is critical. Our findings further suggest that LoRA can match or surpass full fine-tuning, depending on the GeoFM and downstream task.

\subsection{Model Generalization}

We evaluate geographic generalization using geographic hold-out sets (GHOS). First, we compare Prithvi 2.0 300M embeddings after pre-training, full fine-tuning, and LoRA fine-tuning. Figure~\ref{fig:embeddings_sen1floods11} visualizes the Sen1Floods11 embeddings, while reBEN 7k is provided in the supplementary material. The model trained without geographic metadata naturally clusters images by region. The geographic hold-out set from Bolivia forms two clusters after pre-training. Full fine-tuning disrupts this structure, suggesting that it forgets some pre-trained features. At the same time, LoRA better preserves geographic clustering, highlighting its potential to mitigate catastrophic forgetting, a key challenge in fine-tuning foundation models.

\begin{figure}[tbh]
    \centering
    \includegraphics[width=\textwidth]{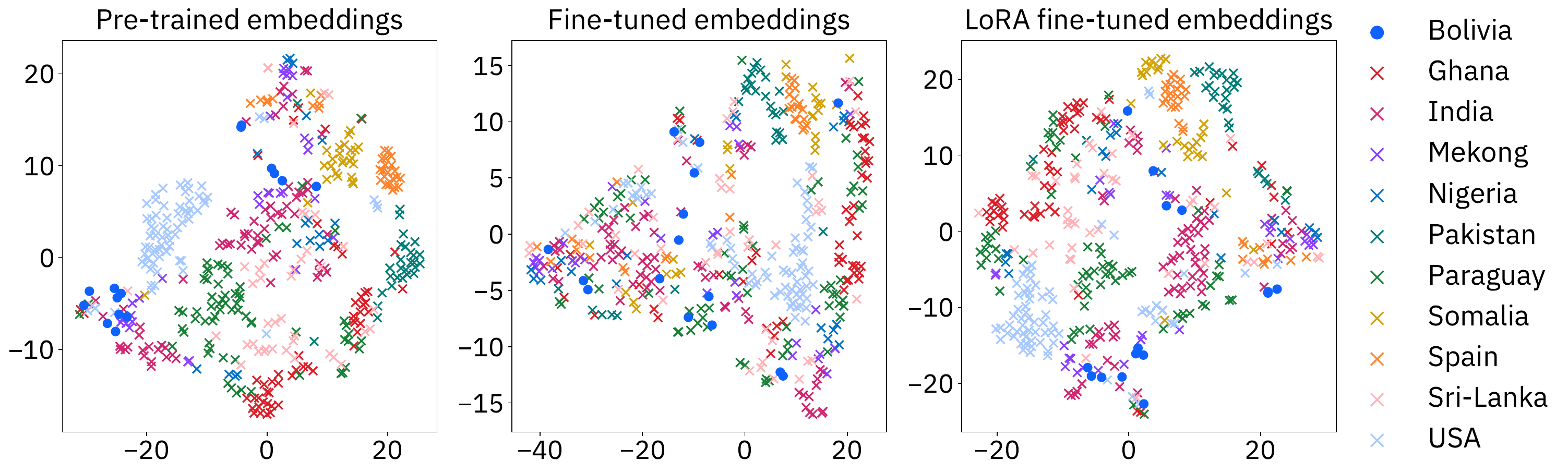}
    \caption{Prithvi 2.0 300M embeddings of the Sen1Floods11 dataset colored by region. We averaged the patch embeddings per image and applied t-SNE.} 
    \label{fig:embeddings_sen1floods11}
\end{figure}

To quantify these patterns, we compute the minimum Euclidean distance of each sample to the training set and average over validation, test, and hold-out splits. The results (see supplementary material) confirm that GHOS samples are significantly farther from training data than validation and test samples, which follow the training distribution. Full fine-tuning reduces distances for all Sen1Floods11 splits by a factor of two but unexpectedly increases them for reBEN~7k. LoRA fine-tuning increases distances for both datasets but maintains the pre-training embedding structure in the t-SNE plot. In all cases, GHOS distances remain higher than those of validation and test sets, highlighting the challenge of geographic generalization.

\begin{table}[tbh]
    \centering
    \caption{Comparison between the in-distribution test set and the geographic hold-out set (GHOS) using mIoU (\%) $\uparrow$, averaged over five runs. We highlight the two best-performing combinations in bold and underlined.}
    \label{tab:ood}
    \setlength{\tabcolsep}{2pt}
    \begin{tabularx}{\linewidth}{llCcCc}
    \toprule
     & & \multicolumn{2}{c}{Sen1Floods11} & \multicolumn{2}{c}{reBEN 7k} \\
    Model & Method & Test mIoU & GHOS mIoU & Test mIoU & GHOS mIoU\\
    \midrule
    UNet (Rand.) & Full FT & \textbf{90.75} & 87.81 & 34.62 & 25.43 \\
\midrule
ViT (IN-21K) & Full FT & 89.19 & 82.67 & 36.15 & 27.14 \\
\midrule
\multirow{2}{*}{DeCUR} & Linear Prob. & 80.83 & 74.76 & 28.29 & 23.29 \\
 & Full FT & 86.87 & 85.84 & 36.05 & 27.60 \\
\midrule
\multirow{4}{*}{Clay v1} & Linear Prob. & 89.57 & 84.88 & 32.65 & 24.38 \\
 & VPT & 89.67 & 87.29 & 36.87 & 28.77 \\
& LoRA & \underline{90.41} & \textbf{88.88} & \underline{38.67} & 28.44 \\
 & Full FT & \underline{90.41} & \underline{88.33} & \underline{38.67} & 29.07 \\
\midrule
\multirow{5}{*}{\shortstack[l]{Prithvi 1.0\\100M}} & Linear Prob. & 88.78 & 79.57 & 27.07 & 24.04 \\
 & VPT & 89.03 & 64.35 & 29.98 & 22.65 \\
& LoRA & 89.33 & 61.25 & 30.60 & 24.09 \\
 & ViT Adapter & 87.72 & 82.31 & 32.90 & 25.66 \\
 & Full FT & 89.02 & 74.16 & 31.82 & 24.53 \\
\midrule
\multirow{5}{*}{\shortstack[l]{Prithvi 2.0\\300M}} & Linear Prob. & 88.08 & 83.19 & 31.25 & 24.86 \\
 & VPT & 89.31 & 86.19 & 38.40 & \underline{29.83} \\
 & LoRA & 90.04 & 87.57 & \textbf{38.84} & \textbf{30.21} \\
 & ViT Adapter & 88.52 & 84.94 & 35.14 & 26.99 \\
 & Full FT & 90.13 & 82.07 & 37.42 & 28.12 \\
    \bottomrule
    \end{tabularx}
\end{table}

Table~\ref{tab:ood} quantifies the geographic generalization gaps: on average, mIoU drops by 7.79pp for reBEN 7k and 7.31pp for Sen1Floods11 across all settings and models. DeCUR exhibits the smallest drop with full fine-tuning (-4.74pp), while Prithvi 1.0 experiences the largest with LoRA (-17.30pp). The best-performing models across both in-distribution and GHOS sets are Prithvi 2.0 300M with LoRA, Clay (full fine-tuning), and Clay with LoRA. Notably, LoRA significantly improves geographic generalization for Prithvi 2.0 300M but has minimal impact for Clay.  
These results highlight that geographic generalization remains an open challenge for GeoFMs. However, they outperform a randomly initialized UNet (+2.2pp on GHOS) and an ImageNet-pretrained ViT (+3.7pp), with LoRA proving particularly beneficial for generalization which is in line with results from literature~\cite{bafghi2024parameter}.
That being said, PEFT methods may struggle in scenarios with extreme distribution shifts, such as across sensor modalities (e.g., from optical to SAR) or when dealing with very high-resolution imagery. In such cases, pre-trained features are expected to transfer poorly, and PEFT's restricted capacity to adapt the encoder weights could become a bottleneck~\cite{bafghi2025fine}. 

To assess model robustness to varying input bands, we conducted experiments with DeCUR and Clay using only six spectral bands (Prithvi bands) under both linear probing and full fine-tuning (results in the supplementary material). Reducing input channels resulted in a minor performance drop, typically within 1-2pp. No significant difference was observed between frozen encoder models and fully fine-tuned ones, suggesting that GeoFMs can adapt to missing bands without updating early layers. The performance loss is therefore attributed to the missing spectral information rather than to poor model adaptation.

\subsection{Decoder Architecture}

\begin{figure}[tbh]
    \centering
    \includegraphics[width=0.7\textwidth]{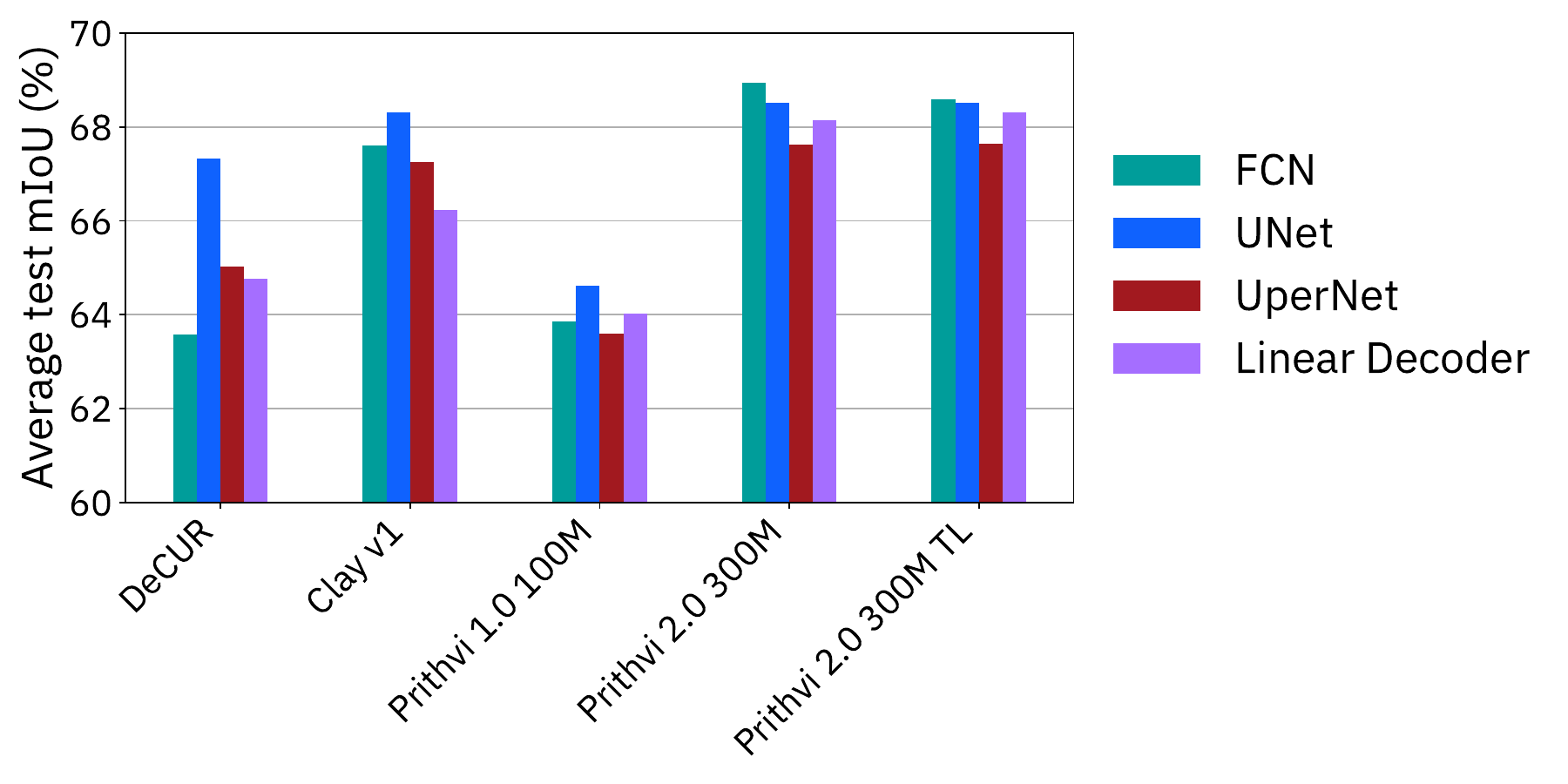}
    \caption{Average test mIoU (\%) $\uparrow$ over five datasets for each model and decoder combination with five runs each.} 
    \label{fig:decoders}
\end{figure}

Figure~\ref{fig:decoders} compares different decoder architectures across all evaluated foundation models using full fine-tuning. We also include Prithvi 2.0 300M TL, which incorporates additional metadata for reference.
The decoder performance varies across GeoFMs, but the UNet decoder consistently achieves strong results, ranking within the top two across all architectures. FCN performs best for Prithvi 2.0 but lags behind UNet for other models. UperNet, despite having a similar parameter count, underperforms across most settings.

\begin{figure}[tbh]
    \centering
    \includegraphics[width=\textwidth]{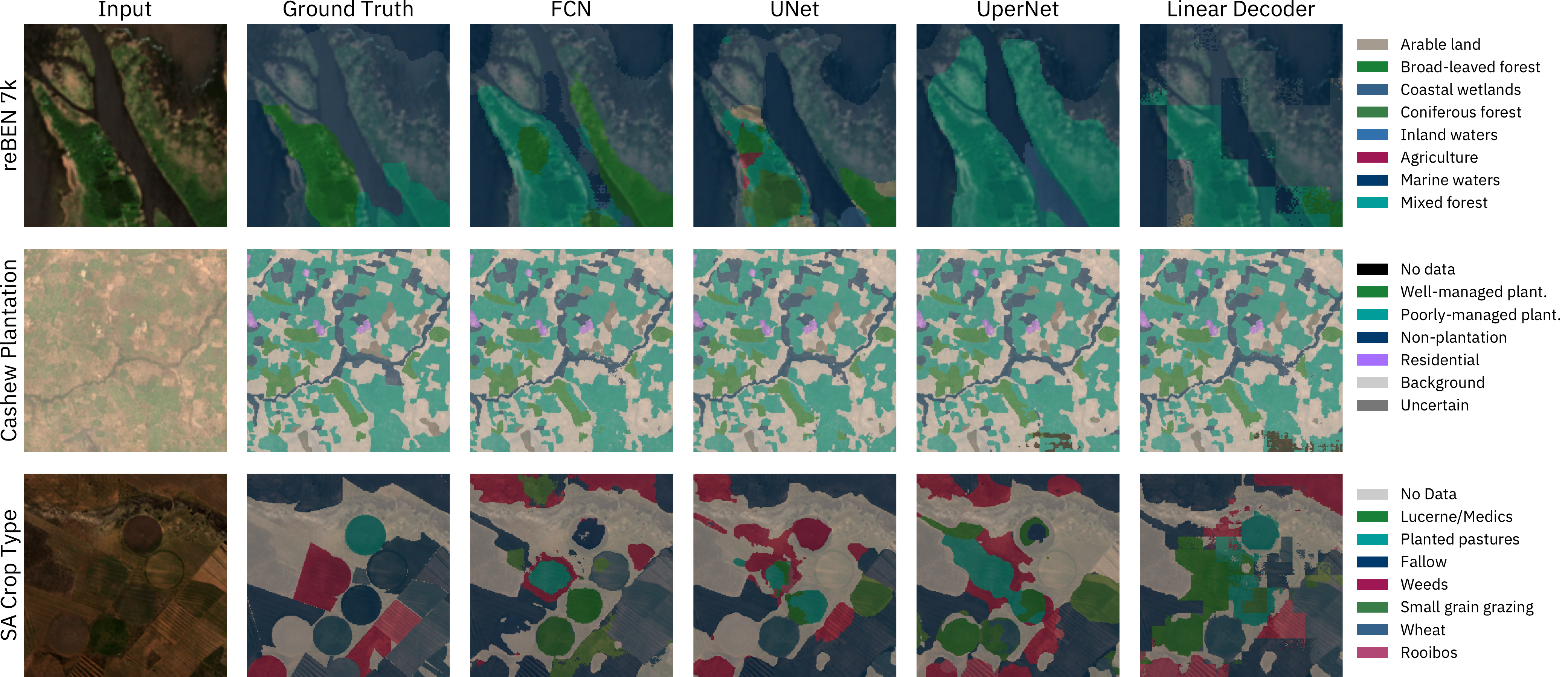}
    \caption{Predictions of Prithvi 2.0 300M with different decoders and unfrozen backbone. The examples are selected to showcase patchy predictions from FCN and the linear decoder, while UperNet and UNet results are smoother.} 
    \label{fig:predictions}
\end{figure}

Surprisingly, with only a single layer, the linear decoder performs competitively and even outperforms UperNet for Prithvi models. However, this result is primarily influenced by a 5pp drop in UperNet's performance on the Cashew Plantation dataset. Combined with our findings from frozen encoder experiments, it suggests that GeoFMs can effectively learn non-linear functions within the encoder, reducing reliance on complex decoders for performance comparisons. However, the lower capacity of the linear decoder cannot capture small features, as shown in Figure~\ref{fig:predictions}, and is unsuitable for applications.

While FCN achieves a high performance, it has a major drawback: Visual analysis of predictions reveals that FCN (and the linear decoder) produces patchy outputs, whereas UNet and UperNet generate more spatially consistent results (see Figure~\ref{fig:predictions}). This is likely because UNet and UperNet access multi-scale feature maps, allowing them to preserve low-level spatial information, which helps with fine-grained segmentation. The visualized images are selected to show failures and do not fully represent the overall prediction quality. The UNet decoder leads to the best predictions, considering performance and visual quality.

\section{Conclusion}

We have extensively evaluated PEFT techniques for geospatial foundation models, analyzing different fine-tuning strategies, geographic generalization, and decoder architectures. Our results show that LoRA matches or exceeds full fine-tuning performance while significantly reducing memory and potentially training time, making it the most effective PEFT method. The decoder comparisons highlight UNet as the best-performing architecture, while metadata has minimal impact on fine-tuning. Geographic generalization remains an open challenge, though GeoFMs outperform standard baselines, especially when fine-tuned with LoRA. These findings suggest LoRA as a suitable fine-tuning strategy for EO applications, particularly in resource-constrained settings with large datasets. However, our study is limited to three models, and PEFT effectiveness may vary with architecture and data. Future work should explore more models and tasks, and assess additional PEFT variants like TEA~\cite{hu2024tea} or DEFLECT~\cite{deflect2025} to identify robust strategies. All models and techniques are available in TerraTorch, enabling scalable adaptation for real-world use.


%
%
%
\newpage

\clearpage
\setcounter{page}{1}

\title{Fine-tune Smarter, Not Harder: \\Parameter-Efficient Fine-Tuning for \\Geospatial Foundation Models}

\titlerunning{Parameter-Efficient Fine-Tuning for Geospatial Foundation Models}


\author{Marti Escofet, F., Blumenstiel, B., Scheibenreif, L., Fraccaro, P., Schindler, K.}

\authorrunning{Marti Escofet, F., Blumenstiel, B., Scheibenreif, L., Fraccaro, P. et al.}

\institute{\large{Supplementary Material}}

\maketitle

We provide additional information for the datasets and experiments. Specifically, we list additional results with PEFT methods, for model generalization, the decoder architecture, and experiments with metadata. 

\section{Datasets}

\begin{figure}
    \centering
    \includegraphics[width=\linewidth]{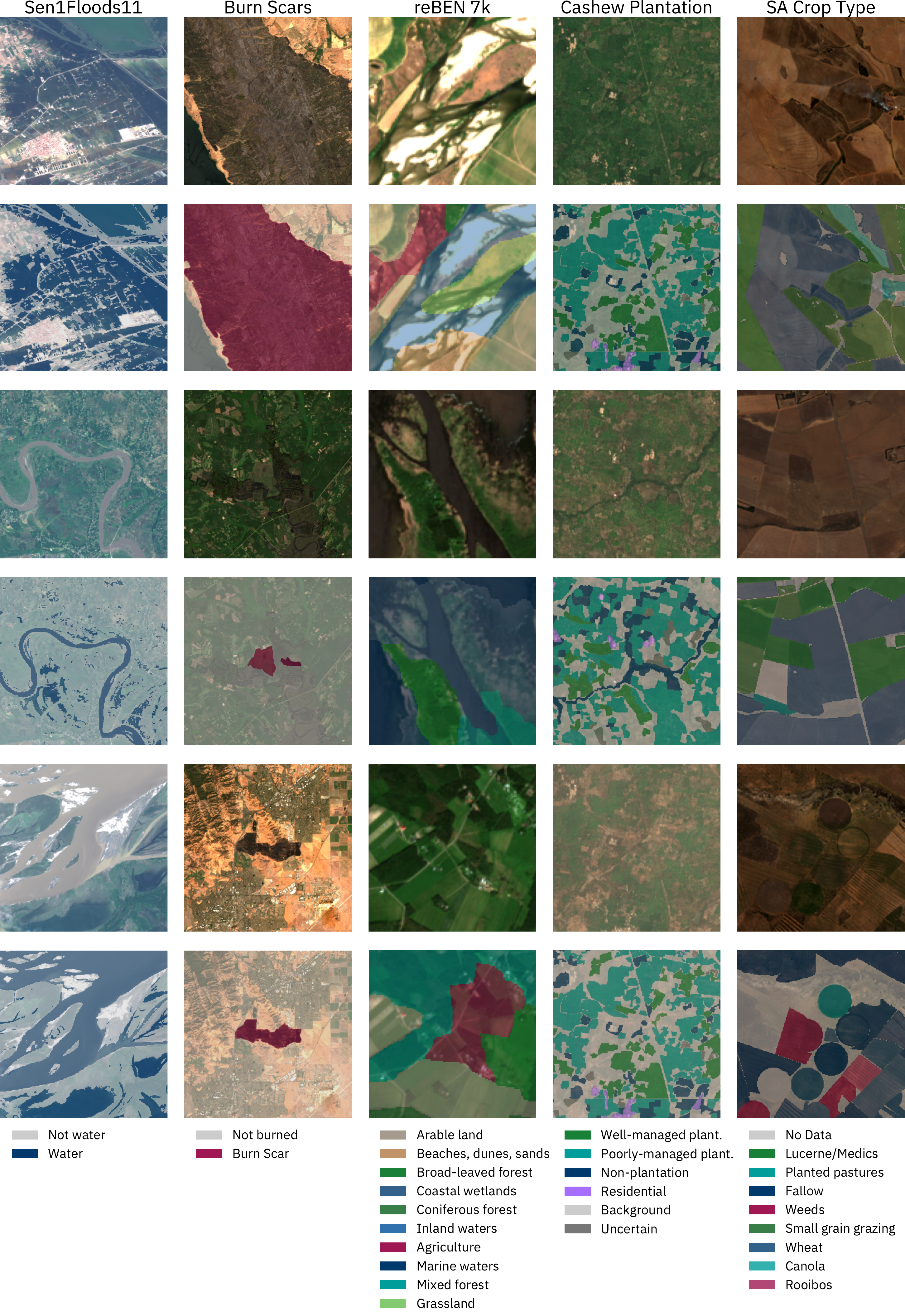}
    \caption{Examples from the downstream datasets and their ground truth annotations. We only list classes that are visual in the examples.}
    \label{fig:examples_sup}
\end{figure}

We provide three samples and their annotated masks for each dataset in Figure~\ref{fig:examples_sup} to show the diversity of the data, the quality of the annotations, and the variations in appearance across different samples.

\section{Additional results}

This section includes additional results for PEFT methods, model generalization, decoder tests, and metadata experiments.

\subsection{PEFT}
The main results section presents the average values for the PEFT experiment. However, for completeness, this appendix includes the same table with standard deviations (Table~\ref{tab:peft_supplementary}) to provide additional context regarding variability. Most standard deviations are below 0.6pp, with some exceptions for the Cashew Plantation dataset. This phenomenon occurs because the test set of the Cashew Plantation dataset comprises only 50 samples from two locations, resulting in highly similar labels. Consequently, even minor variations in the model's predictions can lead to significant fluctuations in test performance, contributing to the higher standard deviations observed.

\begin{table}[tbh]
    \centering
    \scriptsize
    \caption{Test mIoU (\%) $\uparrow$ for the evaluated models and PEFT techniques, with mean and standard deviation computed over five runs.}
    \label{tab:peft_supplementary}
    \setlength{\tabcolsep}{2pt}
    \begin{tabularx}{\linewidth}{llCCCCC}
    \toprule
    Model & Method & Sen1F11 & Burn Scars & reBEN 7k & Cashew Plant. & SACrop \\
    \midrule
    UNet (Rand.) & Full FT & \textbf{90.75} $\pm$ 0.5 & 89.34 $\pm$ 1.0 & 34.62 $\pm$ 0.6 & \textbf{81.81} $\pm$ 6.7 & 34.99 $\pm$ 0.8 \\
\midrule
ViT (IN-21K) & Full FT & 89.19 $\pm$ 0.3 & 92.31 $\pm$ 0.6 & 36.15 $\pm$ 0.3 & 80.05 $\pm$ 0.4 & 37.98 $\pm$ 0.5 \\
\midrule
\multirow{2}{*}{DeCUR} & Linear Prob. & 80.83 $\pm$ 0.2 & 78.17 $\pm$ 0.3 & 28.29 $\pm$ 0.1 & 16.05 $\pm$ 0.3 & 20.93 $\pm$ 0.3 \\
 & Full FT & 86.87 $\pm$ 0.2 & 89.48 $\pm$ 0.8 & 36.05 $\pm$ 0.3 & 79.59 $\pm$ 0.1 & 34.21 $\pm$ 0.4 \\
 \midrule
\multirow{4}{*}{Clay v1} & Linear Prob. & 89.57 $\pm$ 0.0 & 83.17 $\pm$ 0.1 & 32.65 $\pm$ 0.1 & 27.50 $\pm$ 0.1 & 31.10 $\pm$ 0.1 \\
 & VPT & 89.67 $\pm$ 0.3 & 90.67 $\pm$ 0.5 & 36.87 $\pm$ 0.2 & 28.38 $\pm$ 0.2 & 36.89 $\pm$ 0.5 \\
& LoRA & \underline{90.41} $\pm$ 0.2 & 92.74 $\pm$ 0.2 & \underline{38.67} $\pm$ 0.5 & 62.29 $\pm$ 1.5 & 39.64 $\pm$ 0.6 \\
 & Full FT & \underline{90.41} $\pm$ 0.2 & 91.58 $\pm$ 0.4 & \underline{38.67} $\pm$ 0.4 & 72.03 $\pm$ 1.0 & \underline{40.22} $\pm$ 0.6 \\
\midrule
\multirow{5}{*}{Prithvi 1.0 100M} & Linear Prob. & 88.78 $\pm$ 0.0 & 83.23 $\pm$ 0.0 & 27.07 $\pm$ 0.1 & 25.33 $\pm$ 0.2 & 26.06 $\pm$ 0.0 \\
& VPT & 89.03 $\pm$ 0.1 & 85.17 $\pm$ 0.8 & 29.98 $\pm$ 0.4 & 29.24 $\pm$ 0.5 & 29.62 $\pm$ 0.4 \\
& LoRA & 89.33 $\pm$ 0.1 & 89.34 $\pm$ 0.3 & 30.60 $\pm$ 0.5 & 53.00 $\pm$ 2.0 & 31.58 $\pm$ 1.2 \\
 & ViT Adapter & 87.72 $\pm$ 0.3 & 89.47 $\pm$ 0.6 & 32.90 $\pm$ 0.4 & 73.31 $\pm$ 0.8 & 32.34 $\pm$ 0.3 \\
 & Full FT & 89.02 $\pm$ 0.1 & 89.31 $\pm$ 0.4 & 31.82 $\pm$ 0.5 & 77.41 $\pm$ 1.4 & 32.59 $\pm$ 0.4 \\
\midrule
\multirow{5}{*}{Prithvi 2.0 300M} & Linear Prob. & 88.08 $\pm$ 0.1 & 83.90 $\pm$ 0.4 & 31.25 $\pm$ 0.1 & 27.29 $\pm$ 0.4 & 27.26 $\pm$ 0.2 \\
& VPT & 89.31 $\pm$ 0.2 & 92.16 $\pm$ 0.4 & 38.40 $\pm$ 0.3 & 62.00 $\pm$ 2.6 & 39.33 $\pm$ 0.4 \\
& LoRA & 90.04 $\pm$ 0.2 & \textbf{93.33} $\pm$ 0.6 & \textbf{38.84} $\pm$ 0.2 & 77.53 $\pm$ 0.5 & \textbf{40.35} $\pm$ 0.4 \\
 & ViT Adapter & 88.52 $\pm$ 0.3 & 91.95 $\pm$ 0.4 & 35.14 $\pm$ 0.5 & 75.92 $\pm$ 0.3 & 37.24 $\pm$ 0.5 \\
 & Full FT & 90.13 $\pm$ 0.2 & \underline{92.85} $\pm$ 0.8 & 37.42 $\pm$ 0.6 & \underline{80.58} $\pm$ 0.4 & 39.74 $\pm$ 0.3 \\
    \bottomrule
    \end{tabularx}
\end{table}

Additionally, Table~\ref{tab:peft_memory} compares the GPU memory usage of various PEFT techniques with Clay v1 and Prithvi 2.0 300M across datasets. As expected, full fine-tuning results in the highest memory consumption for both models, with Prithvi 2.0 300M reaching up to 21.1 GB on some datasets. In contrast, linear probing is the most memory-efficient, using less than 5 GB in all cases—which represents ~23\% of full fine-tuning memory for both models. Among PEFT methods, LoRA offers a favorable trade-off: it achieves strong performance while maintaining significantly lower memory usage than full fine-tuning, at around 73–93\%. ViT Adapter and VPT fall in between, with ViT Adapter consuming more memory than LoRA but slightly less than full FT on Prithvi. Notably, the memory overhead of each method is consistent across datasets, with larger architectures like Prithvi showing greater absolute savings. These results highlight LoRA's advantage in settings where memory constraints are a limiting factor, offering a scalable alternative to full fine-tuning.

\begin{table}[htb]
    \centering
    \scriptsize
    \caption{Memory usage (in GB) across different PEFT methods for Clay v1 and Prithvi 2.0 300M on five datasets. The final column reports the average memory usage and its percentage relative to full fine-tuning (FT) for each model.}
    \label{tab:peft_memory}
    \setlength{\tabcolsep}{2pt}
    \begin{tabularx}{\linewidth}{llCCCCCc}
    \toprule
    Model & Method & Sen1F11 & Burn Scars & reBEN 7k & Cashew Plant. & SACrop & Mean (perc.)\\
    \midrule
\multirow{4}{*}{Clay v1} & Linear Prob. & 4.8 & 4.6 & 2.6 & 2.6 & 2.6 & 3.5 (22.6\%) \\
 & VPT & 17.0 & 16.9 & 7.1 & 6.0 & 6.0 & 10.6 (69.2\%) \\
 & LoRA & 24.1 & 23.9 & 7.8 & 7.6 & 7.6 & 14.2 (92.8\%) \\
 & Full FT & 24.9 & 24.9 & 9.0 & 8.9 & 8.9 & 15.3 (100.0\%) \\
\midrule
\multirow{5}{*}{\shortstack[l]{Prithvi 2.0\\300M}} & Linear Prob. & 4.1 & 4.1 & 3.3 & 3.1 & 3.2 & 3.6 (23.7\%) \\
 & VPT & 13.1 & 13.1 & 8.9 & 6.2 & 6.3 & 9.5 (63.1\%) \\
 & LoRA & 17.3 & 17.3 & 6.9 & 6.7 & 6.8 & 11.0 (72.8\%) \\
 & ViT Adapter & 20.0 & 20.0 & 7.3 & 7.4 & 7.2 & 12.4 (81.9\%) \\
 & Full FT & 21.1 & 21.1 & 11.2 & 10.9 & 11.0 & 15.1 (100.0\%) \\
    \bottomrule
    \end{tabularx}
\end{table}

\subsection{Model Generalization}

We analyze the embedding space of Prithvi 2.0 300M and distances between different splits in Table~\ref{tab:distances}.
Therefore, we compute the minimum Euclidean distance of each sample to the training set and average it over validation, test, and hold-out splits in the Sen1Floods11 and rebEN 7k datasets. This shows that the distances of the geographic hold-out set remain higher than those of the validation and test sets, highlighting the challenge of geographic generalization.

\begin{table}[tbh]
    \centering
    \caption{Minimum Euclidean distance to the training samples in the embedding space of Prithvi 2.0 300M, averaged over all images of each split. We report the distances after pre-training (PT), full fine-tuning (FT), and LoRA fine-tuning.}
    \label{tab:distances}
    \setlength{\tabcolsep}{4pt}
    \begin{tabular}{lcccccc}
    \toprule
     & \multicolumn{3}{c}{Sen1Floods11} & \multicolumn{3}{c}{reBEN 7k} \\
    Split & PT & Full FT & LoRA & PT & FT & LoRA \\
    \midrule
    Validation & 3.92 & 2.03 & 4.14 & 3.17 & 7.10 & 8.92 \\
    Test & 4.03 & 2.08 & 4.13 & 3.19 & 7.22 & 9.17 \\
    Geo. hold-out & 6.41 & 2.98 & 6.87 & 3.64 & 7.70 & 10.05 \\
    \bottomrule
    \end{tabular}
\end{table}
\begin{table}[htb]
    \centering
    \scriptsize
    \caption{Comparison between the in-distribution test set and the geographic hold-out set (GHOS) using mIoU (\%) $\uparrow$, with mean and standard deviation computed over five runs.}
    \label{tab:ood_supplementary}
    \setlength{\tabcolsep}{2pt}
    \begin{tabularx}{\linewidth}{llCcCc}
    \toprule
     & & \multicolumn{2}{c}{Sen1Floods11} & \multicolumn{2}{c}{reBEN 7k} \\
    Model & Method & Test mIoU & GHOS mIoU & Test mIoU & GHOS mIoU\\
    \midrule
    UNet (Rand.) & Full FT & \textbf{90.75} $\pm$ 0.5 & 87.81 $\pm$ 0.4 & 34.62 $\pm$ 0.6 & 25.43 $\pm$ 0.7 \\
\midrule
ViT (IN-21K) & Full FT & 89.19 $\pm$ 0.3 & 82.67 $\pm$ 2.4 & 36.15 $\pm$ 0.3 & 27.14 $\pm$ 0.5 \\
\midrule
\multirow{2}{*}{DeCUR} & Linear Prob. & 80.83 $\pm$ 0.2 & 74.76 $\pm$ 0.6 & 28.29 $\pm$ 0.1 & 23.29 $\pm$ 0.1 \\
 & Full FT & 86.87 $\pm$ 0.2 & 85.84 $\pm$ 0.4 & 36.05 $\pm$ 0.3 & 27.60 $\pm$ 0.3 \\
 \midrule
\multirow{4}{*}{Clay v1}  & Linear Prob. & 89.57 $\pm$ 0.0 & 84.88 $\pm$ 0.0 & 32.65 $\pm$ 0.1 & 24.38 $\pm$ 0.1 \\
 & VPT & 89.67 $\pm$ 0.3 & 87.29 $\pm$ 1.2 & 36.87 $\pm$ 0.2 & 28.77 $\pm$ 0.5 \\
& LoRA & \underline{90.41} $\pm$ 0.2 & \textbf{88.88} $\pm$ 0.6 & \underline{38.67} $\pm$ 0.5 & 28.44 $\pm$ 0.4 \\
 & Full FT & \underline{90.41} $\pm$ 0.2 & \underline{88.33} $\pm$ 0.5 & \underline{38.67} $\pm$ 0.4 & 29.07 $\pm$ 2.0 \\
\midrule
\multirow{5}{*}{Prithvi 1.0 100M}  & Linear Prob. & 88.78 $\pm$ 0.0 & 79.57 $\pm$ 0.3 & 27.07 $\pm$ 0.1 & 24.04 $\pm$ 0.2 \\
 & VPT & 89.03 $\pm$ 0.1 & 64.35 $\pm$ 9.0 & 29.98 $\pm$ 0.4 & 22.65 $\pm$ 0.4 \\
& LoRA & 89.33 $\pm$ 0.1 & 61.25 $\pm$ 4.6 & 30.60 $\pm$ 0.5 & 24.09 $\pm$ 0.7 \\
 & ViT Adapter & 87.72 $\pm$ 0.3 & 82.31 $\pm$ 1.4 & 32.90 $\pm$ 0.4 & 25.66 $\pm$ 0.7 \\
 & Full FT & 89.02 $\pm$ 0.1 & 74.16 $\pm$ 0.8 & 31.82 $\pm$ 0.5 & 24.53 $\pm$ 0.5 \\
\midrule
\multirow{5}{*}{Prithvi 2.0 300M}  & Linear Prob. & 88.08 $\pm$ 0.1 & 83.19 $\pm$ 0.6 & 31.25 $\pm$ 0.1 & 24.86 $\pm$ 0.2 \\
 & VPT & 89.31 $\pm$ 0.2 & 86.19 $\pm$ 0.9 & 38.40 $\pm$ 0.3 & \underline{29.83} $\pm$ 0.9 \\
& LoRA & 90.04 $\pm$ 0.2 & 87.57 $\pm$ 0.8 & \textbf{38.84} $\pm$ 0.2 & \textbf{30.21} $\pm$ 0.7 \\
 & ViT Adapter & 88.52 $\pm$ 0.3 & 84.94 $\pm$ 0.5 & 35.14 $\pm$ 0.5 & 26.99 $\pm$ 0.8 \\
 & Full FT & 90.13 $\pm$ 0.2 & 82.07 $\pm$ 2.0 & 37.42 $\pm$ 0.6 & 28.12 $\pm$ 0.6 \\
    \bottomrule
    \end{tabularx}
\end{table}

We also include the corresponding table with standard deviations for the GHOS experiments (see Table~\ref{tab:ood_supplementary}). Notably, the standard deviations for GHOS mIoU tend to be larger than those in the test set, likely due to the increased variability and distribution shifts inherent in geographic generalization.

\begin{figure}
    \centering
    \includegraphics[width=\textwidth]{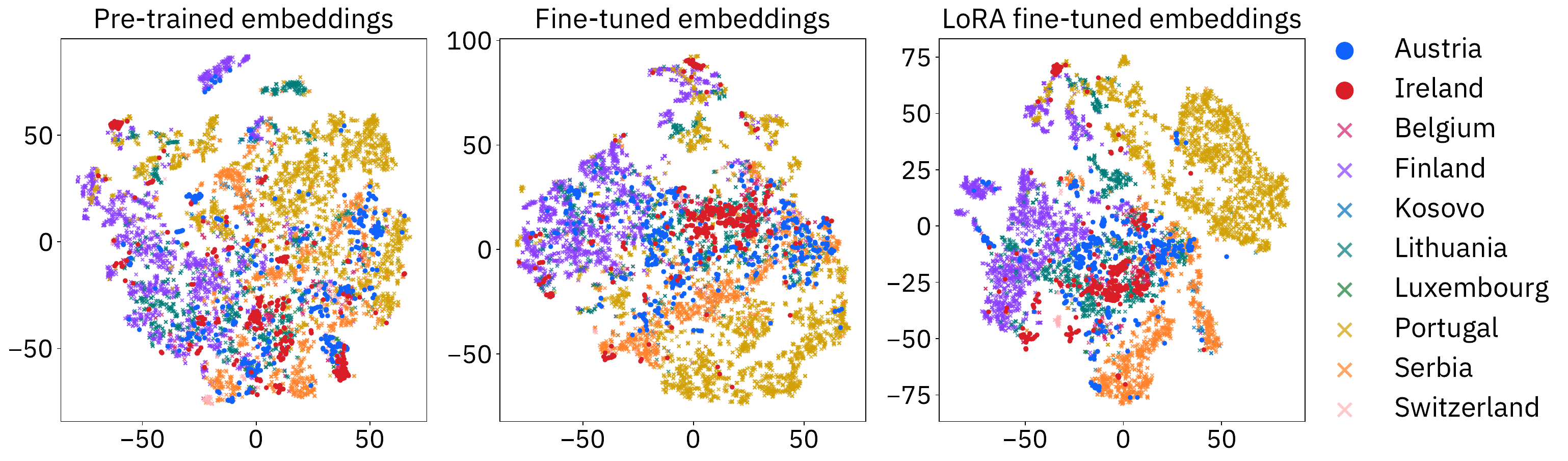}
    \caption{Prithvi 2.0 300M embeddings of the reBEN 7k dataset colored by region. We averaged the patch embeddings per image and applied t-SNE.} 
    \label{fig:embeddings_reben}
\end{figure}

\begin{figure}
    \centering
    \includegraphics[width=\textwidth]{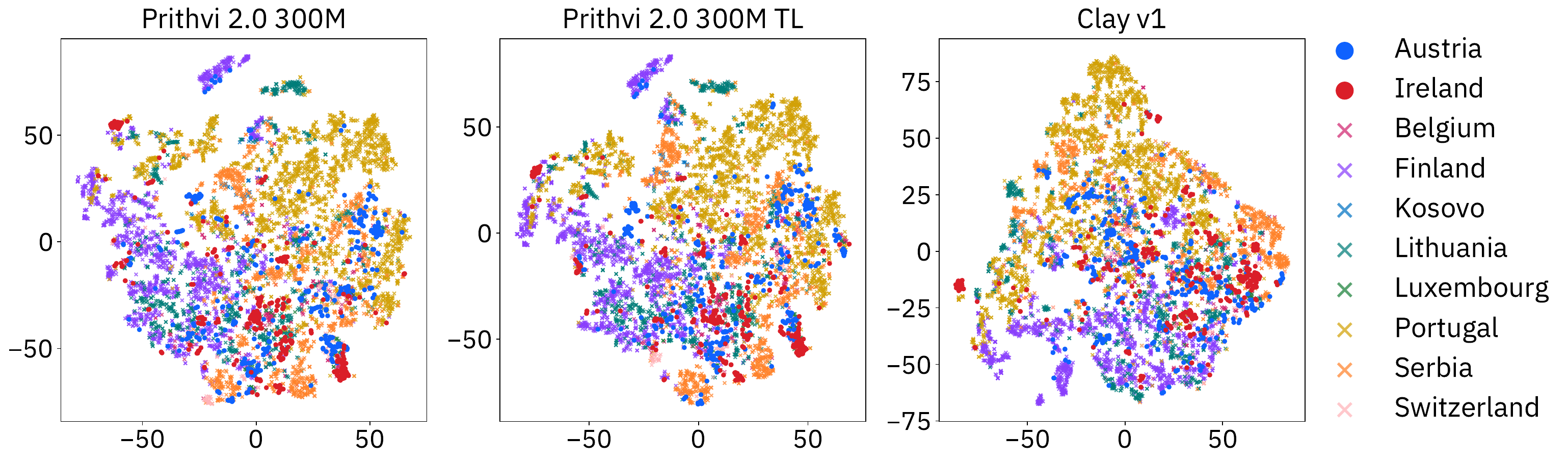}
    \caption{Pre-trained embeddings of Prithvi 2.0 300M, Prithvi 2.0 300M TL, and Clay v1 for the reBEN 7k dataset colored by region. We averaged the patch embeddings per image and applied t-SNE.} 
    \label{fig:embeddings_reben_tl}
\end{figure}

\begin{figure}
    \centering
    \includegraphics[width=\textwidth]{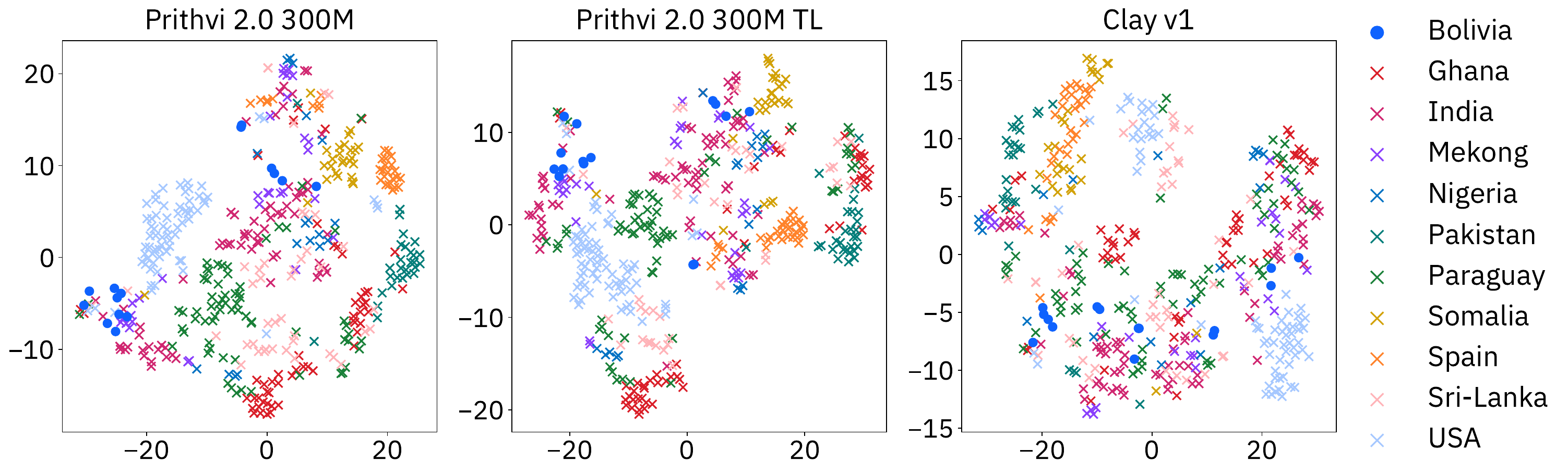}
    \caption{Pre-trained embeddings of Prithvi 2.0 300M, Prithvi 2.0 300M TL, and Clay v1 for the Sen1Floods11 dataset colored by region. We averaged the patch embeddings per image and applied t-SNE.} 
    \label{fig:embeddings_sen1floods11_tl}
\end{figure}

In addition, we provide visualizations of the embeddings for the reBEN 7k dataset with the Prithvi 2.0 300M model in Figure~\ref{fig:embeddings_reben}. Additionally, we compare the pre-trained embeddings of Prithvi 2.0 300M to Prithvi 2.0 300M TL and Clay v1, both pre-trained with metadata, in Figure~\ref{fig:embeddings_reben_tl} and~\ref{fig:embeddings_sen1floods11_tl}.

\subsection{Input Generalization}

\begin{table}[htb]
    \centering
    \caption{Test mIoU (\%) $\uparrow$ for DeCUR and Clay, using different sets of input bands, averaged over five runs. Specifically, we compare the pre-training bands to the six bands used by Prithvi. We provide the delta in parentheses.
    }
    \label{tab:input_bands}
    \setlength{\tabcolsep}{2pt}
    \begin{tabularx}{\linewidth}{lllCCC}
    \toprule
    Model & Method & Input bands & reBEN 7k & Cashew Plant. & SACrop \\
    \midrule
    \multirow{4}{*}{DeCUR} & \multirow{2}{*}{Linear Prob.} & 12 bands & 28.29 & 16.05 & 20.93  \\
    & & 6 bands & 25.93 ($-$2.36 $\downarrow$) & 16.04 ($-$0.01 $\downarrow$) & 19.55 ($-$1.38 $\downarrow$) \\
    & \multirow{2}{*}{Full FT} & 12 bands & 36.05 & 79.59 & 34.21 \\
    & & 6 bands & 34.96 ($-$1.09 $\downarrow$) & 79.69 (+0.10 $\uparrow$) & 32.70 ($-$1.51 $\downarrow$) \\

    \midrule    
    \multirow{6}{*}{Clay v1} & \multirow{2}{*}{Linear Prob.} & 10 bands & 32.65 & 27.50 & 31.10 \\
    & & 6 bands & 31.51 ($-$1.14 $\downarrow$) & 26.84 ($-$0.66 $\downarrow$) & 30.04 ($-$1.06 $\downarrow$) \\
    & \multirow{2}{*}{LoRA} & 10 bands & 38.67 & 62.29 & 39.64 \\
    &  & 6 bands & 37.00 ($-$1.67 $\downarrow$) & 59.78 ($-$2.51 $\downarrow$) & 39.05 ($-$0.59 $\downarrow$) \\

    & \multirow{2}{*}{Full FT} & 10 bands & 38.67 & 72.03 & 40.22 \\
    &  & 6 bands & 37.63 ($-$1.04 $\downarrow$) & 71.12 ($-$0.91 $\downarrow$) & 39.64 ($-$0.58 $\downarrow$) \\
    \bottomrule
    \end{tabularx}
\end{table}

To assess robustness to missing input bands, we test models using only the six pre-training bands of Prithvi (Blue, Green, Red, Narrow NIR, SWIR 1, SWIR 2) and compare against the pre-training bands of DeCUR and Clay. 
Due to normalization, dropping input bands mathematically equals inputting the mean channel value. Therefore, we expect a performance drop when the first encoder layers cannot adapt due to freezing.

Table~\ref{tab:input_bands} shows that the impact of missing bands varies by dataset and model. While Cashew Plantation remains unaffected for DeCUR, other models and dataset combinations mostly see a drop between 1pp and 2pp. Notably, the performance deltas remain consistent across full fine-tuning and linear probing, suggesting that GeoFMs can easily adapt to missing input bands. This implies that observed performance losses stem from the missing spectral information rather than poor model adaptation.

\subsection{Metadata}

\begin{table}[tbh]
    \centering
    \caption{Effect of metadata in pre-training (PT) and fine-tuning (FT) on the test mIoU (\%) $\uparrow$, averaged over five runs. The metadata includes the timestamp and location.}
    \label{tab:metadata}
    \setlength{\tabcolsep}{2pt}
    \begin{tabularx}{\linewidth}{lccCCCCc}
    \toprule
    Model & PT & FT & Sen1F11 & Burn Scars & reBEN 7k & Cashew Plant. & Mean \\
    \midrule
    \multirow{2}{*}{Clay v1} & \multirow{2}{*}{\ding{51}} & \ding{55} & 90.41 & 91.58 & 38.67 & 72.03 & 66.58 \\
     &  & \ding{51} & 90.49 & 91.47 & 38.95 & 72.30 & 66.73 (+0.15 $\uparrow$) \\
    \midrule
    \multirow{3}{*}{Prithvi 2.0 300M} & \ding{55} & \ding{55} & 90.13 & 92.85 & 37.42 & 80.58 & 68.14 \\
    \cmidrule{2-8}
     & \multirow{2}{*}{\ding{51}} & \ding{55} & 90.08 & 92.98 & 37.44 & 79.87 & 67.97 ($-$0.17 $\downarrow$) \\
     &  & \ding{51} & 89.94 & 93.16 & 38.35 & 80.63 & 68.32 (+0.18 $\uparrow$) \\
    \bottomrule
    \end{tabularx}
\end{table}

We investigate the impact of incorporating metadata, specifically temporal and location information, into the model inputs. These experiments use a linear decoder with an unfrozen backbone. The SA~Crop Type dataset was excluded as it lacks metadata, while Cashew includes only temporal metadata.

Clay was pre-trained with metadata, whereas Prithvi 2.0 has two variants: one with metadata (300M TL) and one without (300M). Our experiments show only minimal differences between models trained with and without metadata. Pre-training Prithvi 2.0 with metadata results in a marginally lower average mIoU (-0.17pp), but this advantage is reversed by +0.35pp when metadata is also used during fine-tuning. In comparison, Clay v1 performs slightly better when using metadata in both pre-training and fine-tuning.

The limited impact of metadata suggests that models can infer temporal and geographic patterns directly from image data. This is further supported by Figure~\ref{fig:embeddings_reben}, which visualizes Prithvi 2.0 300M embeddings (trained without metadata) for the reBEN 7k dataset. The apparent clustering in these pre-trained embeddings indicates that the model naturally captures geographic structures without explicit metadata. Thus, adding metadata may provide little additional benefit for downstream tasks.

\clearpage

\subsection{Decoder Architecture}

\begin{figure}
    \centering
    \includegraphics[width=\linewidth]{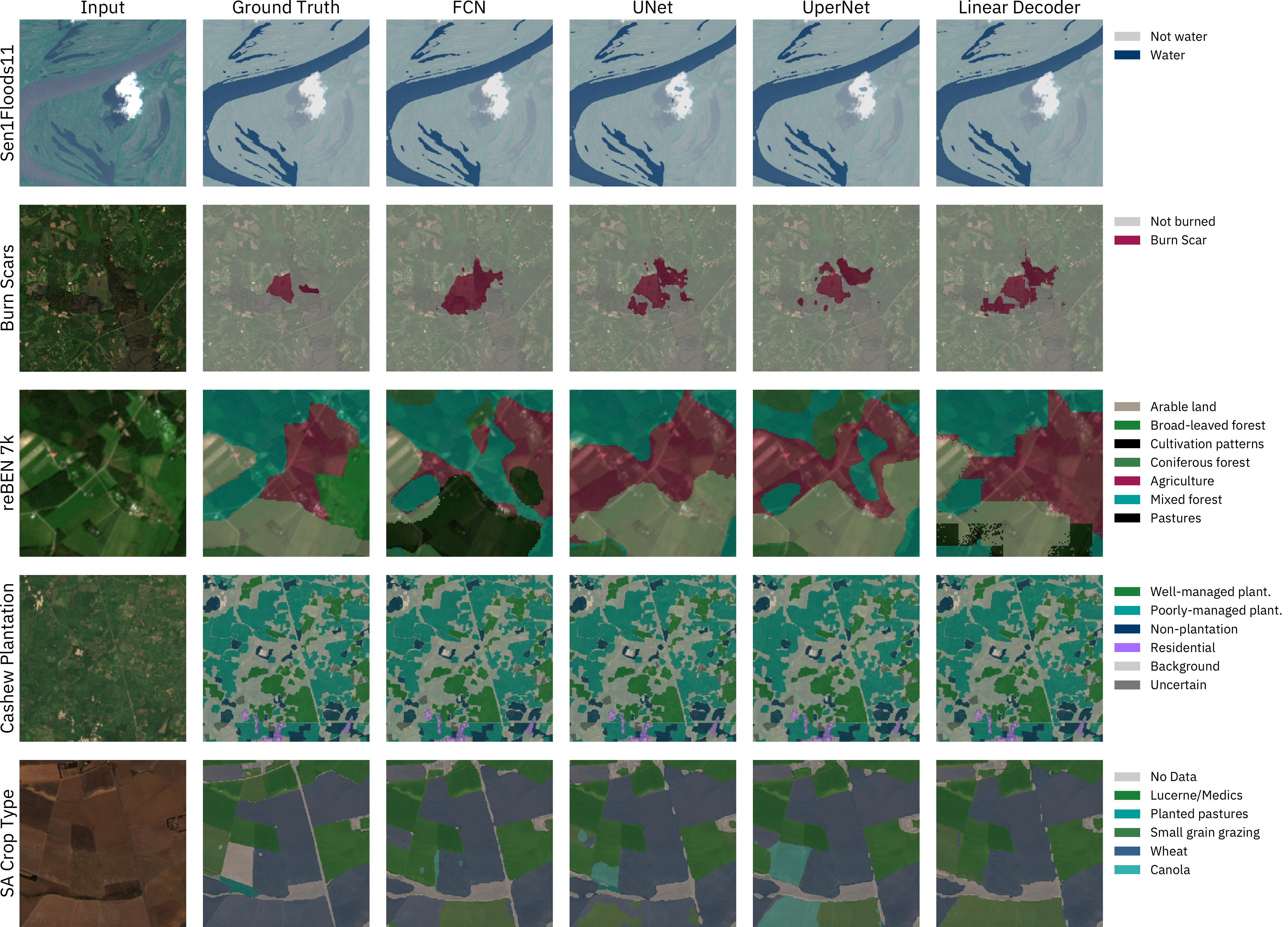}
    \caption{Model predictions of Prithvi 2.0 300M using different decoders.}
    \label{fig:predictions_all}
\end{figure}

We provide more prediction examples of Prithvi 2.0 300M with different decoders in Figure~\ref{fig:predictions_all}.

\end{document}